# Exploration of Summarization by Generative Language Models to Enhance Automated Scoring of Long Essays

Haowei Hua

Princeton University, Princeton, New Jersey, United States

Hong Jiao

University of Maryland, College Park, Maryland, United States

Xinyi Wang

University of Maryland, College Park, Maryland, United States

Corresponding author: Haowei Hua

Email address: jack.hua@princeton.edu

Country affiliation for all three authors: United States

## Abstract

BERT and its variants are extensively explored for automated scoring. However, a limit of 512 tokens for these encoder-based models showed the deficiency in automated scoring of long essays. Thus, this research explores generative language models for automated scoring of long essays via summarization and prompting. The results revealed great improvement of scoring accuracy with QWK increased from 0.822 to 0.8878 for the Learning Agency Lab Automated Essay Scoring 2.0 dataset.

## Background

Encoder-based small language models (SLMs) including BERT (Delvin et al., 2018), deBERTa (He et al., 2020), RoBERTa (Liu et al., 2019) and other BERT family models (Clark et al., 2020; Sasaki & Masada, 2022) have demonstrated great potential in automated scoring (e. g., Fernandez et al., 2023; Jiao, Lnu, & Zhai, 2024; Lotridge, 2023; Shermis & Wilson, 2024). Typically, transformer-based models like BERT and its variants have a maximum token limit of 512. In many practical automated essay scoring, the average essay token length could be more than 512 tokens. A recent Kaggle competition on automated essay scoring (Crossley et al., 2024) includes essays even longer than 3,000 words. Wang et al (2025) compared the performance of BERT and several of its variants with truncation to the maximum token size to 512, and found that feature-based modeling built upon Light Gradient Boosting Machine (LightGBM; Ke et al., 2017) performed the best. However, the exploration conducted by Wang et al. (2025) is far beyond comprehensive.

To process long text more effectively, researchers have developed specialized transformer architectures such as BigBird, Hierarchical BERT (Kong et al., 2022; Zaheer et al., 2021; Zhang et al., 2019), and Longformer (Beltagy et al., 2020). These models incorporate techniques like sparse attention and hierarchical encoding to better manage extended input sequences. Longformer, for instance, combines local windowed attention with selective global attention and supports input lengths up to 4,096 tokens. Its advanced variant, the Longformer Encoder-Decoder (LED), extends this capacity to 16,384 tokens (AI2, 2020/2024; Pang et al., 2022). In addition, generative large language models (LLMs) such as GPT (OpenAI et al., 2024), Claude (Anthropic, 2024), and Gemini (Gemini Team et al., 2024) exhibit even stronger capabilities for processing long texts. GPT-4.1 supports up to 1 million tokens, enabling comprehensive understanding of lengthy and complex documents, a substantial improvement over GPT-4o, which maxed out at 128,000 tokens. Claude 3.7 models can handle up to 200,000 tokens, while Gemini 2.5 Pro can accommodate up to 1,048,576 tokens (roughly 750,000 words or 1,500 pages). These expanded capacities support full-document processing without truncation, making generative LLMs especially valuable for automated long essay scoring.

Recent studies have explored text summarization as a strategic solution to the input-length constraint inherent in BERT-based models (Chowdhury et al., 2024; Gavalan et al., 2024), as summarization has shown to reliably capture essential semantic information (Latif & Wood, 2009; Mutasodirin & Prasojo, 2021; Syahra et al., 2022). Mutasodirin and Prasojo (2021) found that extractive summarization achieved comparable or better F1 performance than most truncation strategies, suggesting its effectiveness as a text-shortening approach for Transformer-based models. Gavalan et al. (2024) demonstrated that applying text summarization as a preprocessing step in BERT-based models significantly improved performance in depression detection tasks.

Wang et al. (2025) evaluated feature-based and BERT-based models for automated long essay scoring with truncation. Truncation may omit key semantic information. Thus, this study seeks to address this limitation by investigating whether LLMs' summarization of long essays can improve automated essay scoring accuracy. Three LLMs including GPT-5, GPT-5 nano, GPT-5 mini, were explored for summarization. These approaches were compared against the truncation-based best-performing language model and feature-based methods previously reported by Wang et al. (2025).

## Methods

### Data

This study utilized the Learning Agency Lab Automated Essay Scoring 2.0 dataset (Crossley et al., 2024), consisting of 24,000 argumentative essays, including 17,307 essays in the training set (Kaggle.com, 2024). This Kaggle competition only made the training data available. Thus, the training dataset was divided into three subsets: training, validation, and test. The sample sizes for these subsets were 13,845, 1,731, and 1,731 essays, respectively as used in Wang et al (2025) for direct comparison. Essay lengths vary from 150 to 1,656 words. Specifically, 4,203 essays have 512 words or fewer, 2,969 essays contain more than 1,000 words, and 360 essays exceed 1,500 words.

Figure 1 presents the distribution of essay word counts in the training dataset, showing a distinctly right-skewed pattern in which the majority of essays fall between 150 and 500 words. The frequency peaks around 250–300 words and gradually declines as essay length increases. Only a small proportion of submissions exceed 800 words, and essay lengthy above 1,000 words are less frequent. This distribution suggests that most responses are concise and within the typical range expected for automated essay scoring tasks. However, the extended right tail indicates the presence of a few unusually long essays, which may require length normalization or summarization to maintain uniformity in model input size and prevent training bias toward longer texts.

Figure 1. Essay Length Distribution for the Original Essays

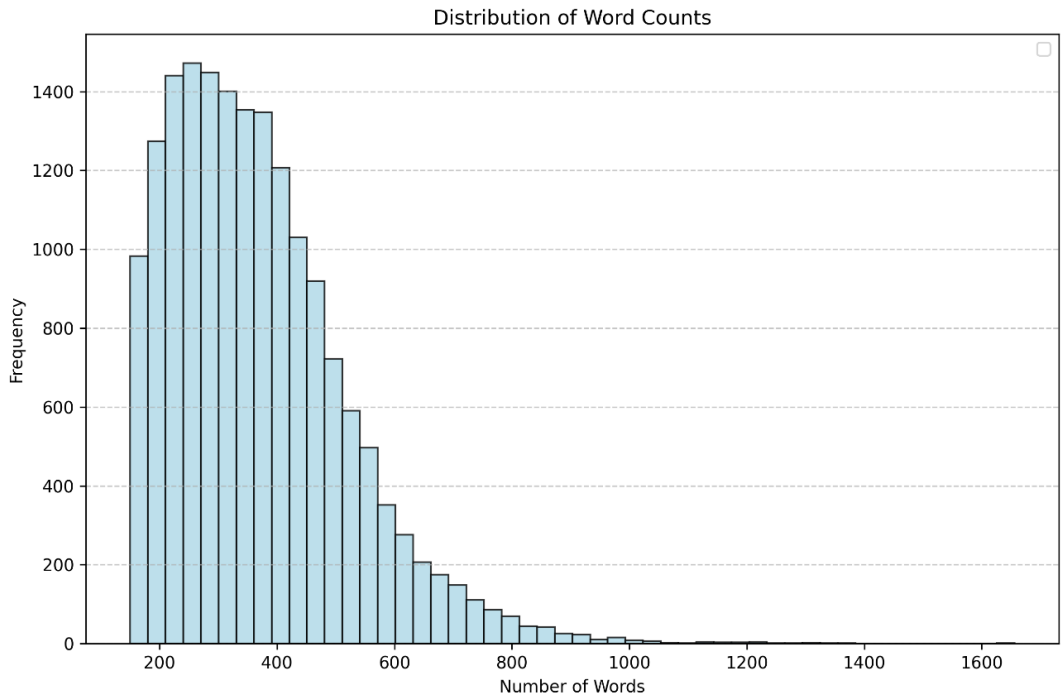

Figure 2. Essay Token Length Distribution for the Original Essays

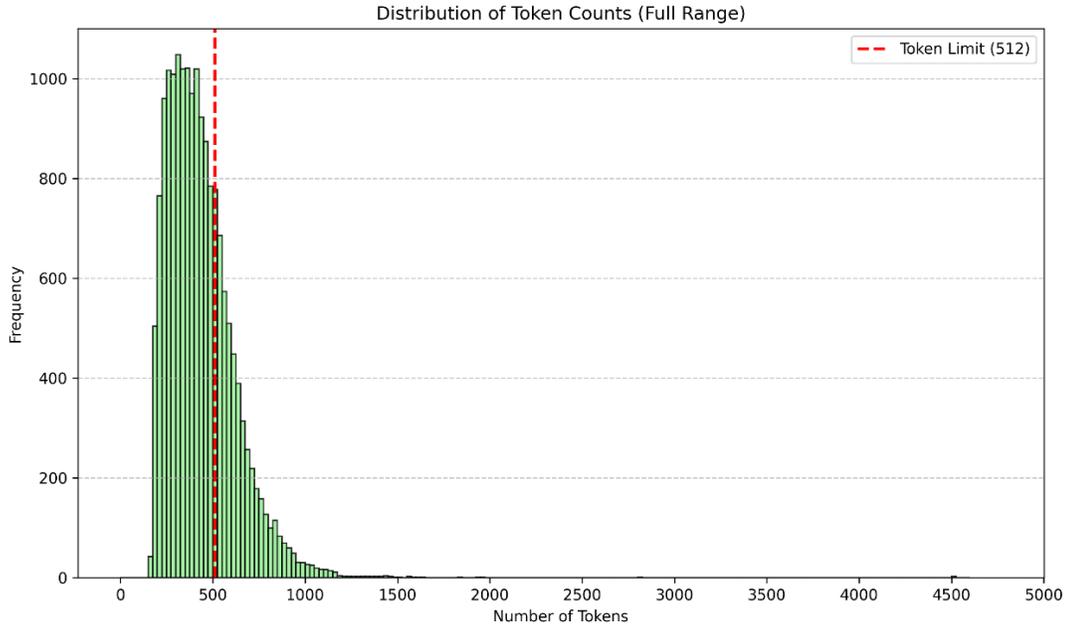

To quantify the textual length of each essay in a manner consistent with large language model (LLM) processing, token counts were computed using the tiktoken library developed by OpenAI. Specifically, the `cl100k_base` encoding was employed, which serves as the standard tokenizer for GPT-4 and GPT-3.5 families. This tokenizer segments text into subword units (*tokens*) rather than whole words, offering a more precise measure of the actual input length that an LLM would interpret during inference. The approach ensures comparability between raw essays and summarized versions in subsequent analyses of model input size and efficiency (OpenAI, 2023).

Figure 2 illustrates the distribution of token counts in the dataset, revealing a strongly right-skewed pattern where most essays cluster between 200 and 500 tokens. The red dashed line marks the 512-token threshold commonly used as the standard limit for model input. Although the majority of essays fall below this boundary, a noticeable tail extends beyond 512 tokens, with a small proportion exceeding 1,000 tokens. This indicates that while most samples are concise enough for direct model input, a subset still surpasses the limit and may require summarization or truncation. The distribution highlights the need for adaptive preprocessing to ensure consistent token-level normalization and efficient training performance.

Figure 3. Essay Token Length Distribution for Essays with 1000 or More Tokens

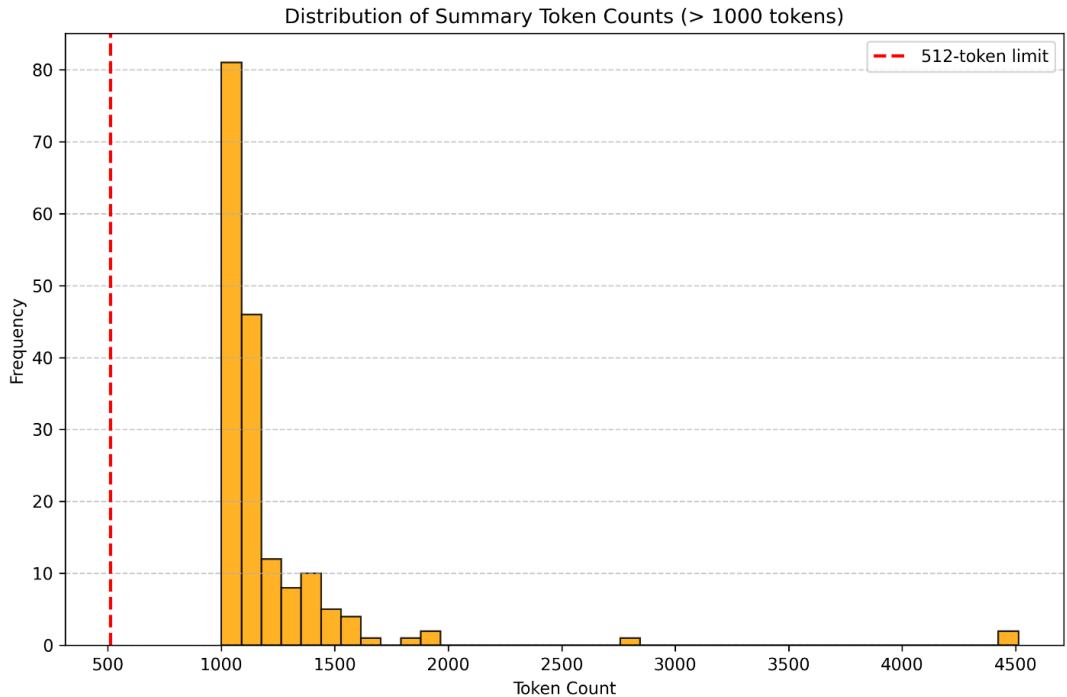

Figure 3 illustrates the distribution of overlength summaries in the dataset, specifically those exceeding 1,000 tokens. The histogram shows that while most of these long samples cluster around 1,000–1,200 tokens, a small number extend considerably further, reaching up to approximately 4,500 tokens. The red dashed line marks the 512-token threshold typically used as the standard input limit for large language models. Although these extended outputs represent a small fraction of the total dataset, they demonstrate clear variability in summarization performance, where the model occasionally fails to adhere to the intended token constraint. This long-tail distribution highlights the necessity of implementing adaptive summarization control or iterative truncation strategies to ensure uniformity and computational efficiency during training and evaluation.

Essay scores range from 1 to 6. Score distribution is presented in Figure 4. While the middle score of 3 has the highest frequency, the scores at the high end of the scale (score 5 and 6) have relatively lower frequencies. The highest score of 6 has the lowest frequency with 0.9% of the total essays. Even the n-count for score 5 is only 5.6% of the total. This may impose class imbalance issues in training machine learning models.

Figure 4. Score Distribution of the Original Essays

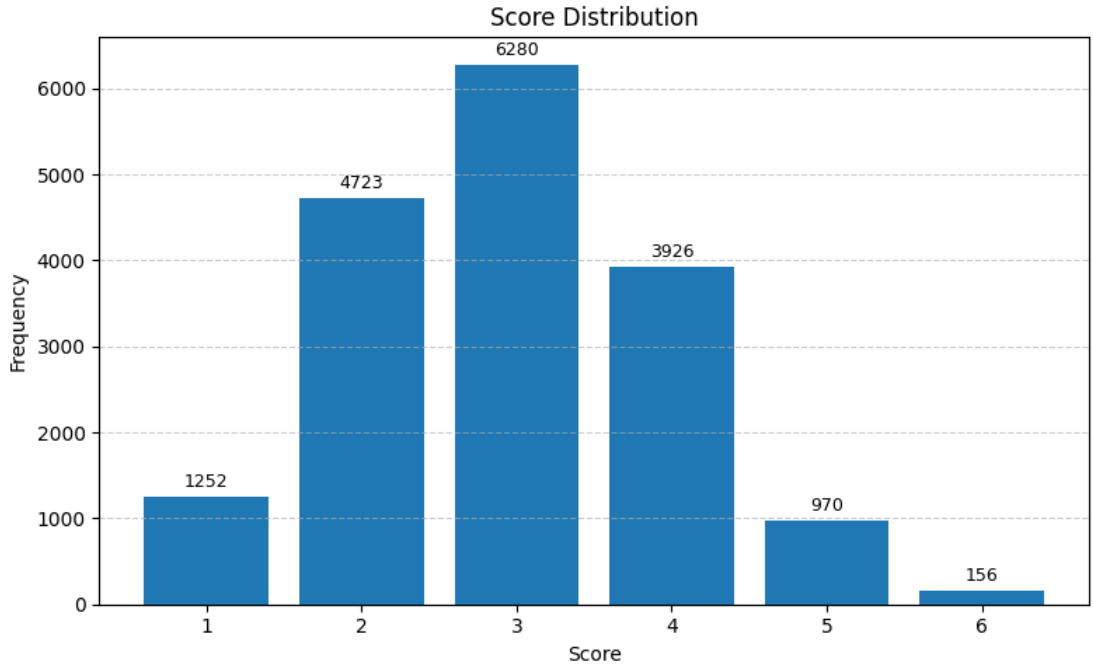

**Summarization by GPT-5**

      As GPT-5 has strong capacity in summarizing text, this study requested GPT-5 to condense the long essays to shorter length via summarization. For summarization, a GPT-based summarization framework was employed to condense long-form text into max-length representations of 512 tokens or fewer, ensuring consistent input dimensions for downstream tasks for automated essay scoring. In the initial stage, all essays were summarized to evaluate the effect of text compression. However, the first-round summarization found some essays still contained over a thousand tokens while some other essays were condensed into way shorter length than 512 tokens which may dramatically led to the loss of valuable linguistic and stylistic features in those essays. To preserve these features, summarization in the second round was applied only when the original text length exceeded 512 tokens, allowing shorter essays to remain in their original form shown in Figure 5.

Figure 5. Illustration of Summarization Process

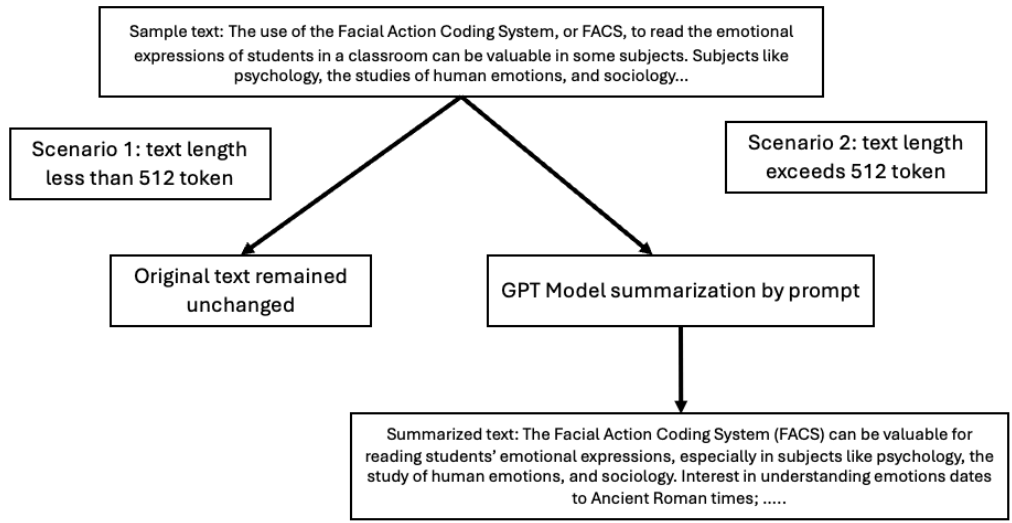

During early experimentation, a length control issue was identified: the GPT-based models (GPT-5, GPT-5-mini, and GPT-5-nano) occasionally produced summaries that exceeded the 512-token limit despite explicit instruction. To mitigate this, an adaptive summarization process was implemented. When a summarized output exceeded the limit, the text was re-summarized with a stricter prompt specifying a target length reduced to 80% of the previous threshold (1000 × 0.8). This iterative process continued until the resulting text met the 512-token constraint, ensuring both compliance with input restrictions and semantic completeness.

All summarization prompts were designed under strict fidelity constraints, preserving original essays' intent, arguments, and terminology while removing redundancy and verbosity through selective compression rather than paraphrasing. Unlike conventional truncation methods, the GPT summarizer leveraged transformer-based contextual reasoning to maintain logical flow and coherence across compressed texts. By enforcing adaptive length control, this summarization process achieved a balance between semantic fidelity and computational efficiency, enabling fair evaluation and model comparison across essays of varying lengths. Prompting used for summarization is provided in Appendix A.

**Models**

This study evaluated the effectiveness of GPT-5 for processing long essays for automated scoring. The first category was the summarization-based methods. For each essay, GPT-5 and its two variants (GPT-5 mini and GPT-5 nano) were used to compress the full text into a shorter version that captures the main ideas. This summary was then used as input to train BERT-based language models (LMs) including BERT-large, deBERTa-base, RoBERTa-base and ELECTRA-base. This method assumes that a LLM generated summary retains the most relevant content for scoring, while reducing input length. These encoder-based language models were trained based on summaries of long essays using GPT-5 mini, GPT-5 nano, and GPT-5 respectively.

This study explored four main modeling approaches for automated essay scoring, each leveraging LLM representations to enhance robustness and interpretability. The first approach involves individual transformer baseline models including BERT, RoBERTa, ELECTRA, and DeBERTa, where each model was fine-tuned separately on three types of summarized datasets by GPT-5, GPT-5-mini, and GPT-5-nano respectively.

Figure 6. Illustration of MLP Neural Network

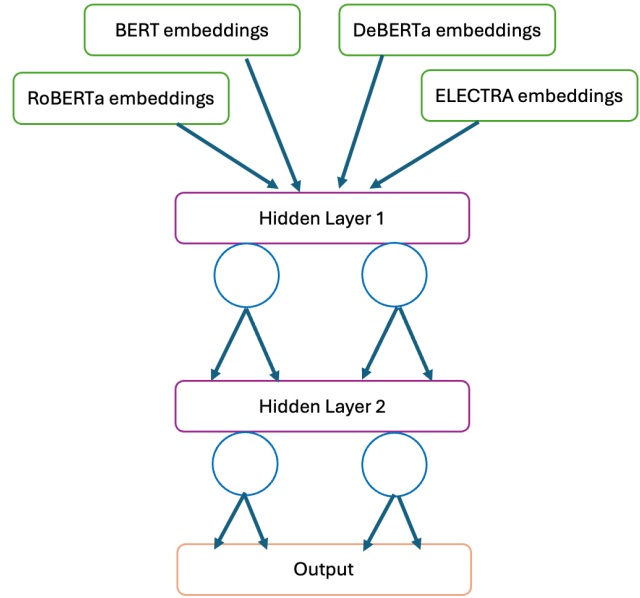

The second approach is embedding-based supervised machine learning model, a Multilayer Perceptron (MLP) in Figure 6. This model concatenated embeddings from all four LMs based on summarized essays with shorter lengths and trained a two-hidden-layer MLP to jointly learn from diverse contextual spaces, capturing deeper inter-model dependencies and complementary linguistic patterns.

Figure 7. Illustration of XGBoost and LightGBM Ensemble Approach

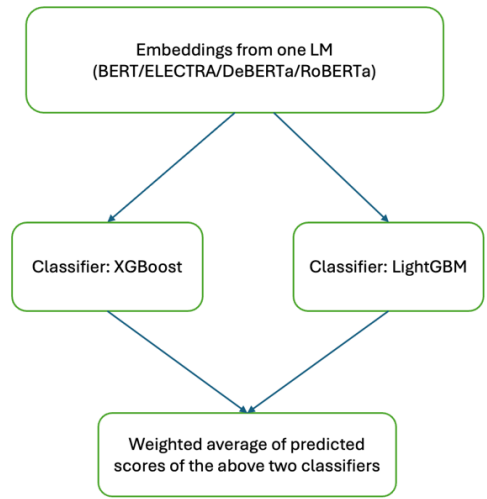

The third approach is ensemble of XGBoost and LightGBM trained on the embeddings from one of the four LMs in Figure 7. This approach integrates the rich contextual embeddings from a single transformer backbone with tree-based gradient boosting algorithms to model complex nonlinear feature interactions. Both XGBoost and LightGBM were trained independently, and their prediction outputs were subsequently combined through a weighted averaging strategy to maximize stability and cross-validation performance. Four ensemble models of XGBoost and LightGBM were trained on a single embedding from BERT, DeBERTa, ELECTRA, and RoBERTa.

Finally, the fourth approach following Yao (2024), the LM-embedding with feature-based Voting ensemble, employed a hard-voting mechanism over five independently trained models that share the same transformer backbone but vary in architecture, regularization, and training objectives. These five models included COPE (Condounder-only Pertuebation Estimator) adversarial training, ordinal regression, ordinal regression with text cleaning, PET (prompt-based example tuning) -style prompt-based training, and multi-scale ordinal regression with MSE loss (Yao, 2024). Each model was trained incorporating handcrafted linguistic features such as paragraph features (length, spelling errors, word/sentence count), sentence features (length, word count), word-level distributions and text vectorization features from TF-IDF Vectorizer and CountVectorizer. These features allow the ensemble to effectively integrate semantic richness from embeddings with explicit structural linguistic indicators of writing quality. Collectively, these approaches reflect a comprehensive exploration of both deep learning and ensemble-based paradigms for improving scoring consistency and generalization across summarized essay datasets.

**Model Performance Evaluation**

To determine how well each method scores summarization of the long essays, Quadratic Weighted Kappa (QWK) was computed as it evaluates how well the model agrees with human raters while accounting for the severity of disagreement. This metric was used in the Kaggle competition.

The QWK score is a statistical measure used to evaluate the agreement between two sets of ordinal ratings—typically between human raters and an automated system (Cohen, 1968). Unlike simple accuracy, QWK takes into account both the degree of disagreement and the ordinal nature of the scoring scale, penalizing larger discrepancies more heavily than smaller ones. It ranges from –1 to 1, where 1 indicates perfect agreement, 0 represents random agreement, and negative values indicate systematic disagreement. Mathematically, the QWK score is defined as:

$$k = 1 - \frac{\sum w_{i,j} O_{i,j}}{\sum w_{i,j} E_{i,j}}, \qquad (1)$$

where $O_{i,j}$ is the observed matrix representing the normalized counts of essays receiving score *i* from one rater and *j* from another, $E_{i,j}$ is the expected matrix assuming random agreement (derived from the outer product of the score histograms), and $W_{i,j}$ is the weight matrix, calculated as:

$$w_{i,j} = \frac{(i-j)^2}{(N-1)^2}, \qquad (2)$$

*N* refers to the number of possible score categories. In the ASAP 2.0 (Automated Student Assessment Prize) dataset, scores are typically assigned on a 1–6 grade scale, where 1 indicates the lowest performance and 6 the highest.

In the context of automated essay scoring (AES)**,** QWK is widely used as the primary evaluation metric because it not only measures how often model predictions match human scores but also how close the model is when it misses. For example, predicting a 5 instead of a 6 incurs a smaller penalty than predicting a 1 instead of a 6. This makes QWK particularly suitable for assessing model consistency with human judgment in ordinal grading systems like ASAP, where maintaining reliable, human-like scoring sensitivity is essential.

**Model Training and Testing**

All models were trained and tested using the same dataset split to ensure fair comparisons. The same test dataset was used for model performance evaluation. The sample sizes for train, validation, and test were 13,845, 1,731, and 1,731 essays, respectively, the same as those used in Wang et al (2025) for direct comparison.

We concatenated the embeddings of multiple models, including BERT, RoBERTa, ELECTRA, and DeBERTa, as an input to a two-hidden-layer MLP. The MLP model was configured with an input dimensionality of 5,376, two hidden layers of 3,200 and 1,600 units respectively, and an output layer of 6 classes. Training employed the cross-entropy loss function with the Adam optimizer, using a learning rate of 0.001, batch size of 128, and 8 epochs. The model is trained using 10-fold cross-validation and evaluated with QWK to save the best model.

For the fourth ensemble, models were trained with 4-fold stratified cross-validation on both subsets, with checkpoints saved at each fold to maximize reproducibility. Several models used ordinal-regression objectives to reflect the ordered nature of essay scores, while others experimented with PET-like (pattern-exploiting) training and hierarchical pooling to enrich representations. To further refine predictions, threshold optimization, an adapted Nelder–Mead search plus a custom loop to fine-tune the final cut point yielding consistent CV gains was adapted. Finally, individual predictions were aggregated with a hard-voting ensemble for improved stability across validation metrics. In the competition setting, incorporated data augmentation was included beyond prompt-preserving transformations within the official corpus (Yao, 2025). This approach was supplemented training with larger, thematically related essay datasets that were outside the Kaggle-provided source. These external samples were rubric-mapped, de-duplicated, and screened for prompt overlap to prevent leakage, then used for pretraining finetuning to boost generalization.

**Results**

**Summarization**

Table 1 presents the descriptive statistics of text length across the original essays and three GPT-based summarization settings. The original training dataset contains substantially longer essays, with an average token count of 436.46 and word count of 368.35, reflecting the high verbosity typical of unprocessed writing samples. In contrast, the GPT-5-mini summaries exhibit strong compression effects, reducing the mean token count to 244.51 and mean word count to 200.33, while maintaining moderate variability (Std = 84.09 for tokens, 69.98 for words). The GPT-5-nano model produces slightly longer outputs on average (Mean = 258.70 tokens, 216.86 words), though its compression remains consistent, indicating stable summarization efficiency under tighter parameter constraints. Finally, the GPT-5 model achieves the most semantically rich summaries, with a mean token count of 273.80 and word count of 222.15, suggesting a deliberate trade-off between brevity and contextual completeness. Overall, the progressive increase in average token and word counts from GPT-5-mini to GPT-5 underscores a hierarchy of summarization depth, where larger models preserve more linguistic and semantic detail while maintaining the 512-token constraint.

Table 1. Summary of Word and Token Counts for Original Essays and Summarized Essays (Training) by GPT-5 mini, GPT-5 nano and GPT-5

| **Original Essays** | **Max** | **Min** | **Mean** | **Std** |
| --- | --- | --- | --- | --- |
| token_count | 4512 | 163 | 436.46 | 185.88 |
| word_count | 1656 | 150 | 368.348 | 150.39 |
| **GPT-5 nano** | | | | |
| token_count | 512 | 163 | 244.51 | 84.09 |
| word_count | 480 | 150 | 200.33 | 69.98 |
| **GPT-5 mini** | | | | |
| token_count | 512 | 163 | 258.7 | 78.66 |
| word_count | 476 | 150 | 216.86 | 64.98 |
| **GPT-5** | | | | |
| token_count | 512 | 163 | 273.8 | 86.63 |
| word_count | 467 | 150 | 222.15 | 68.49 |

Figure 8 displays the distribution of token counts for the GPT-5-mini summarized essays, revealing a unimodal and slightly right-skewed pattern centered around 200–250 tokens. The majority of summarized essays fall well below the 512-token limit (marked by the red dashed line), indicating that the summarization process effectively compressed the original texts while maintaining consistency in length. The smooth decline beyond 300 tokens and the sparse occurrence of samples approaching the upper limit suggest that very few summaries exceeded the intended compression threshold. Overall, this distribution demonstrates that the GPT-5-mini summarizer produced concise and length-stable representations, ensuring efficient model input handling and minimizing the risk of truncation in downstream processing.

Figure 8. Token Length Distribution of Summarized Essays by GPT-5 mini

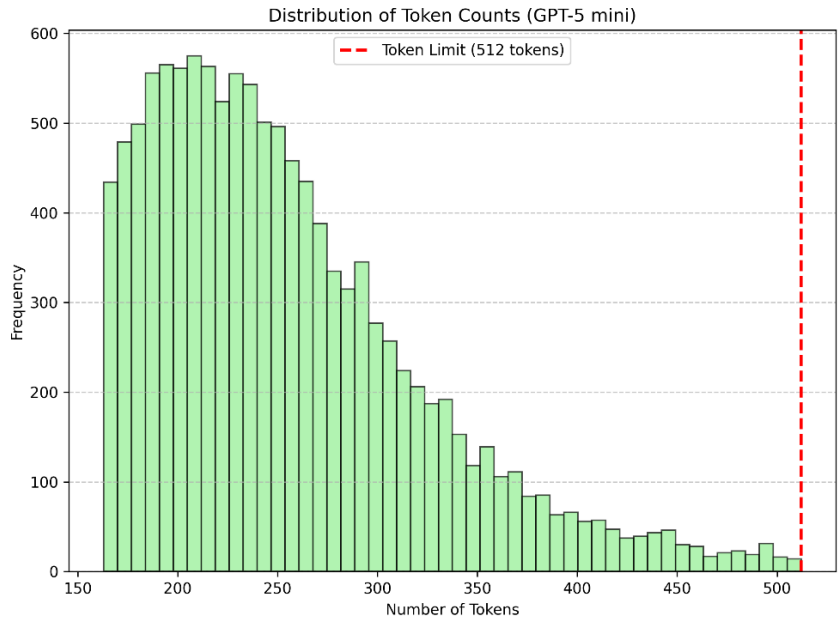

Figure 9 illustrates the distribution of token counts for summarized essays by GPT-5-nano, showing a relatively symmetric peak centered around 250–275 tokens. Compared to GPT-5-mini, the overall distribution shifts slightly toward longer summaries, suggesting that the nano variant generated more text per sample despite operating under the same 512-token constraint. The majority of summaries remained well below the 512-token upper limit (indicated by the red dashed line), demonstrating effective adherence to length control during the summarization process. However, the broader spread and higher mean token count imply that GPT-5-nano tended to produce denser outputs, possibly due to less aggressive compression and a higher tendency to retain contextual details. This pattern reflects a balance between brevity and completeness, though it may introduce variability in downstream model efficiency compared to GPT-5-mini.

Figure 9. Token Length Distribution of Summarized Essays by GPT-5 nano

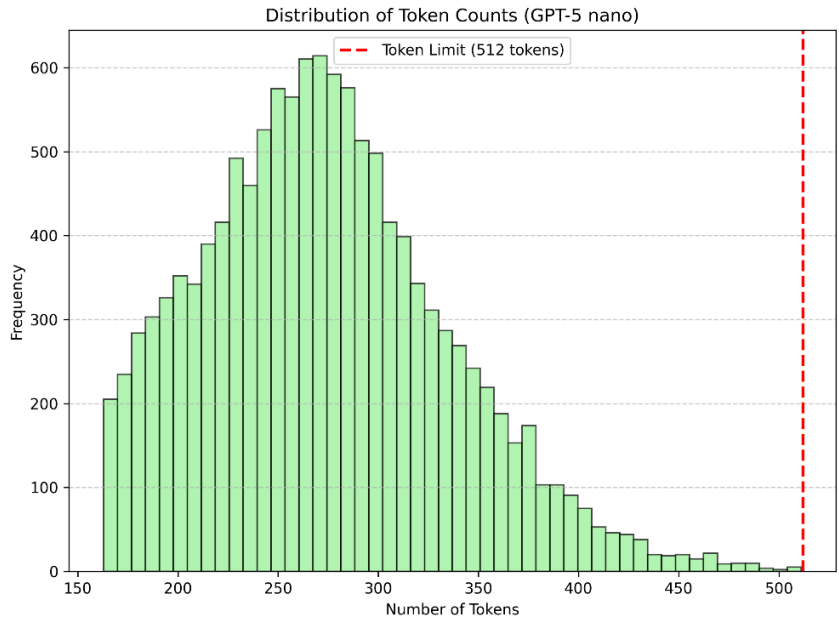

Figure 10. Token Length Distribution of Summarized Essays by GPT-5

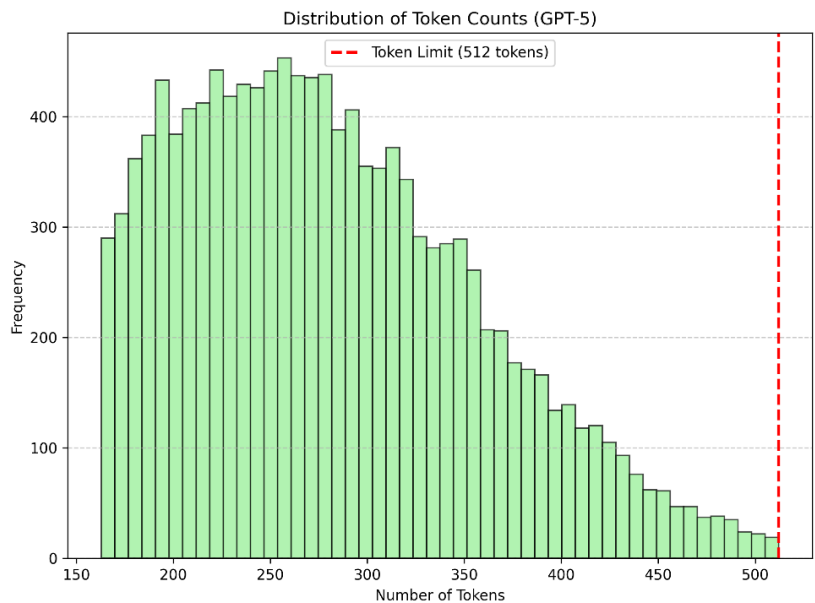

  Figure 10 presents the distribution of token counts for the GPT-5 summarizer, showing a relatively broad and balanced spread centered around 230–280 tokens. Compared with GPT-5-mini and GPT-5-nano, this version produces slightly longer and more evenly distributed summaries, indicating that the model prioritizes semantic completeness and contextual preservation over aggressive compression. Most samples fall below the 512-token limit (marked by the red dashed line), confirming that the summarizer maintains effective control over text length while allowing flexibility for content-rich essays. The smooth decline in frequency beyond 300 tokens and the reduced presence of short outputs reflect GPT-5's adaptive summarization capability, which dynamically adjusts summary length based on content complexity. Overall, this distribution suggests that GPT-5 achieves the best balance between length efficiency and information fidelity among the three summarization settings.

**Model Performance**

Table 2 presents the QWK performance of four language models (BERT, RoBERTa, DeBERTa, and ELECTRA) across the three summarized datasets by GPT-5, GPT-5-nano, and GPT-5-mini. RoBERTa model trained on GPT-5 summarized essays yielded the highest QWK (0.8389) among all models. LMs trained on essays summarized by GPT-5-nano performed better than their counterparts trained on essays summarized by GPT-5 mini. Overall, RoBERTa and DeBERTa based transformer models achieved higher QWK scores, with peaks at 0.813 and 0.8098 respectively on the GPT-5-mini dataset, outperforming BERT (0.8001) and ELECTRA (0.7833) under the same condition. In general, no matter which GPT-5 model was used for essay summarization, RoBERTa and DeBERTa performed better than BERT and ELECTRA.

Table 2. Model Performance of LMs on Essays Summarized by Different LLMs.

| Summarization LLM | LM | QWK | Test File |
|---|---|---|---|
| GPT-5 | BERT | 0.7927 | GPT-5 |
| GPT-5 | roBERTa | 0.8389 | GPT-5 |
| GPT-5 | deBERTa | 0.8169 | GPT-5 |
| GPT-5 | ELECTRA | 0.7965 | GPT-5 |
| GPT-5-mini | BERT | 0.8086 | GPT-5-mini |
| GPT-5-mini | roBERTa | 0.8203 | GPT-5-mini |
| GPT-5-mini | deBERTa | 0.8160 | GPT-5-mini |
| GPT-5-mini | ELECTRA | 0.8136 | GPT-5-mini |
| GPT-5-nano | BERT | 0.8001 | GPT-5-nano |
| GPT-5-nano | roBERTa | 0.8130 | GPT-5-nano |
| GPT-5-nano | deBERTa | 0.8098 | GPT-5-nano |
| GPT-5-nano | ELECTRA | 0.7833 | GPT-5-nano |

Table 3 summarizes the MLP model performance based on concatenated embeddings from four LMs. "GPT-5-ensemble" indicates that the MLP model is based on the training dataset summarized by GPT-5 and embeddings from all four LMs: BERT, RoBERTa, ELECTRA, and DeBERTa. These trained models were tested on different test files summarized by different GPT-5 models. GPT-5 ensemble consistently outperforms the others, achieving the highest QWK of 0.8632 on the GPT-5 test set, 0.8546 on GPT-5-mini, and 0.8456 on GPT-5-nano, highlighting its strong adaptability to varying text granularities. In comparison, the GPT-5-mini ensemble attained QWK values between 0.8221 and 0.8451, while the GPT-5-nano ensemble produced slightly lower but stable results, ranging from 0.8108 to 0.8277. Overall, these findings indicate that ensemble strategies effectively increased essay scoring accuracy, with the GPT-5 ensemble demonstrating the best overall performance and generalization across test conditions. With truncation of the original essays, Wang et al. (2025) achieved QWK of 0.815, lower than the best performing model (QWK of 0.8632) based on GPT-5 summarization.

Table 3. Model Performance of MLP based on Concatenated Embeddings from Four LMs

| Summarization Model | QWK | Test File |
|---|---|---|
| GPT-5 | 0.8456 | GPT-5-nano |
| GPT-5 | 0.8546 | GPT-5-mini |
| <span style="color:red">GPT-5</span> | <span style="color:red">0.8632</span> | <span style="color:red">GPT-5</span> |
| GPT-5-mini | 0.8221 | GPT-5-nano |
| GPT-5-mini | 0.8451 | GPT-5-mini |
| GPT-5-mini | 0.8309 | GPT-5 |
| GPT5-nano | 0.8277 | GPT-5-nano |
| GPT5-nano | 0.8108 | GPT-5-mini |
| GPT5-nano | 0.8204 | GPT-5 |

Table 4. Performance of Ensemble Models of XGBoost and LightGBM Based on Each LM Embedding

| Summarization LLM | LM | QWK | Test File |
|---|---|---|---|
| GPT-5 | BERT | 0.8407 | GPT-5 |
| GPT-5 | roBERTa | 0.8634 | GPT-5 |
| GPT-5 | deBERTa | 0.8402 | GPT-5 |
| GPT-5 | ELECTRA | 0.8387 | GPT-5 |
| GPT-5-mini | BERT | 0.8203 | GPT-5-mini |
| GPT-5-mini | roBERTa | 0.8543 | GPT-5-mini |
| GPT-5-mini | deBERTa | 0.8424 | GPT-5-mini |
| GPT-5-mini | ELECTRA | 0.8502 | GPT-5-mini |
| GPT-5-nano | BERT | 0.8231 | GPT-5-nano |
| GPT-5-nano | roBERTa | 0.8512 | GPT-5-nano |
| GPT-5-nano | deBERTa | 0.8345 | GPT-5-nano |
| GPT-5-nano | ELECTRA | 0.8033 | GPT-5-nano |

Table 4 presents the model performance of ensemble model of XGBoost and LightGBM models based on each LM using summarized essays by each GPT-5 model. "GPT-5-mini BERT" means that the all datasets were summarized by GPT-5 mini and the LM embedding was extracted from BERT. Among all architectures, RoBERTa consistently performed the best, yielding the highest QWK of 0.8634 when GPPT-5 used for essay summarization. DeBERTa generally performed second best except ELECTRA surpassed it when GPT-5 mini was used to summarize essays. The results indicate that RoBERTa and DeBERTa generally captured stronger contextual and structural representations, while the GPT-5-mini and GPT-5-nano datasets maintain competitive performance (both with highest QWK above 0.85) with less computational overhead. The XGBoost and LightGBM ensemble models based on only one single LM (RoBERTa) embeddings proved to be an efficient ensemble approach to achieve comparable QWK as the MLP based on embeddings from four LMs. Wang et al. (2025) reported QWK of 0.783, 0.812, and 0.781 for ensemble of XGBoost and LightBoost based on embeddings from RoBERTa, DeBERTa, and BERT respectively. Our best-performing model based on GPT-5 summarization with roBERTa embeddings yielded way higher QWK (0.8634).

Table 5. Model Performance of LM Embedding and Handcrafted Feature-based Voting

| Model | QWK | Test File |
|---|---|---|
| GPT-5_BERT | 0.8722 | GPT-5 |
| GPT-5_roBERTa | 0.8878 | GPT-5 |
| GPT-5_deBERTa | 0.8798 | GPT-5 |
| GPT-5_ELECTRA | 0.8601 | GPT-5 |
| GPT-5_BERT | 0.8301 | GPT-5-mini |
| GPT-5_roBERTa | 0.8545 | GPT-5-mini |
| GPT-5_deBERTa | 0.8467 | GPT-5-mini |
| GPT-5_ELECTRA | 0.8345 | GPT-5-mini |
| GPT-5_BERT | 0.8318 | GPT-5-nano |
| GPT-5_roBERTa | 0.8506 | GPT-5-nano |
| GPT-5_deBERTa | 0.8427 | GPT-5-nano |
| GPT-5_ELECTRA | 0.8307 | GPT-5-nano |

Given that trained models based on GPT-5 summarized essays always yielded the highest QWK compared with GPT_5 nano and mini, another ensemble model was trained on essays summarized by GPT-5 only. Table 5 summarizes essay scoring accuracy of ensemble models using LM-embedding and handcrafted features in voting ensemble based on essays summarized by GPT-5. "GPT-5_BERT" means that the backbone embedding was obtained from BERT model based on essays summarized by GPT-5. Both embeddings from a single LM and hand-crafted linguistic features were aggregated as input data to train each of the five models introduced above (COPE adversarial training, ordinal regression, ordinal regression with text cleaning, PET-style prompt-based training, and multi-scale ordinal regression with MSE loss). The RoBERTa-based model again achieved the highest overall accuracy, reaching a QWK of 0.8878 on the GPT-5 dataset and maintaining strong generalization on GPT-5-nano (0.8506) and GPT-5-mini (0.8545). DeBERTa followed closely with QWKs of 0.8798 on GPT-5, 0.8467 on GPT-5-mini, and 0.8427 on GPT-5 nano, confirming its robust semantic encoding under ensemble voting. Wang et al. (2025) reported a QWK od 0.822, way lower than our best-performing ensemble model with QWK of 0.8878. Overall, incorporating handcrafted linguistic features such as paragraph length, spelling error rate, and word distributions along with LM embeddings further enhanced the models' scoring accuracy, yielding the best results among all ensemble strategies.

**Summary and Discussion**

This study evaluates GPT-5 summarization in enhancing long essay scoring, extending the work of Wang et al. (2025). Training LMs on essays summarized by GPT-5 improved scoring accuracy compared with the truncation approach reported in Wang et al. (2025), where QWK were all below 0.80 (0.76 for BERT, 0.76 for RoBERTa, 0.769 for DistilBERT, and 0.79 for DeBERTa-v3-large). The empirical evidence provided by this study demonstrated that essays summarized by GPT-5 can be an effective way to increase long essay scoring accuracy.

Overall, the results demonstrate that the GPT-based summarization strategy significantly improves performance on the ASAP dataset, confirming that condensed yet semantically rich representations benefit automated essay scoring. Models trained and tested using the same summarizer version (e.g., GPT-5 with GPT-5, or GPT-5-mini with GPT-5-mini) tend to achieve higher QWK values, suggesting that consistency in summarization granularity helps maintain feature alignment between training and testing distributions. Furthermore, integrating LLM embeddings with traditional machine-learning classifiers such as XGBoost and LightGBM provides additional gains, as these methods effectively capture nonlinear feature relationships from the dense representations. The final voting-based ensemble, which incorporates original text features (e.g., text length, spelling errors, and word distributions) together with embeddings based on summarized essays, achieves the strongest performance overall, indicating that combining raw-text features with compressed semantic representations may represent an optimal approach. However, an important limitation is that portions of this ensemble method rely on non-official, thematically related corpora for auxiliary augmentation. The inclusion of external material makes the approach more tailored to optimizing leaderboard performance than to strict, like-for-like comparability. As a result, reported gains may partially reflect distributional advantages rather than purely architectural merit, and they may be less portable to settings that prohibit external data. Overall, among the summarizers, GPT-5 outperforms both GPT-5-mini and GPT-5-nano, while the relatively weaker results of GPT-5-nano may stem from excessive information loss due to over-compression or limited contextual depth in shorter summaries.

For future work, it will be essential to explore how to balance original-text features and summarized embeddings more effectively, develop ensemble frameworks that integrate both truncated and summarization-based pipelines, and design efficient summarization mechanisms that reduce computational cost while maintaining semantic fidelity.

It is worthy of note that the higher scoring accuracy attained in this current study goes with high cost. Given the high cost of summarization by GPT-5 (with over $1350 cost), it is worth exploring other cost-efficient open-sourced LLMs such as Llama, Mixtral, Phi-3, and GPT-Neo. In addition, other LLMs such as Claude, Grok, DeepSeek, Qwen, and Gemini are worthy of further exploration as well.

**Appendix A**

**Sample prompt used to conduct the summarization task:**

Task: Faithfully compress long essays into a maximum of 512 tokens without improving, embellishing, or altering the author's intent or claims.

Guiding principles:

Fidelity over enhancement: preserve the original thesis, main arguments, structure, tone, and level of certainty. Do not strengthen, weaken, correct, or "polish" the reasoning, style, or rhetoric beyond what is strictly necessary for brevity and coherence.
No new content: do not add interpretations, examples, transitions, or background not present in the original. Do not infer or resolve ambiguities—mirror them.
Preserve key details: retain essential evidence, data, dates, names, definitions, and technical terms; keep figures and claims unchanged.
Maintain voice and stance: keep register, perspective, and nuance; avoid stylistic flourishes not in the source. Only paraphrase to reduce redundancy.
Work step-by-step:

Identify the thesis, primary arguments, and overall structure, noting tone and qualifiers.
Select only the most critical supporting evidence and examples that directly substantiate those arguments.
Rewrite concisely, removing repetition and peripheral material while preserving meaning, emphasis, causality, and logical flow. Do not upgrade clarity by reworking the argument's substance.
Review for fidelity: ensure no new claims were introduced, none were omitted or altered, and qualifiers and uncertainties are intact.
Trim to ≤512 tokens. If further cuts are needed, remove lower-priority details before affecting core arguments, evidence, tone, or stance.
Output format:

Output only the compressed essay text, in coherent natural language.
No meta commentary, explanations, or lists—just the compressed essay.
One paragraph or a few clear paragraphs, totaling no more than 512 tokens.
Do not add titles or summaries beyond what the original conveys.
Example:
Input (excerpt):
[Full essay text about the impact of climate change on global agriculture—several paragraphs, over 2,000 tokens.]

Output:
The effects of climate change on global agriculture are profound, posing risks to food security and rural livelihoods. Rising temperatures and shifting weather patterns threaten crop yields, with some regions experiencing drought while others face floods. Adaptation strategies include developing drought-resistant crops and improving irrigation efficiency. Without cohesive global action to mitigate emissions and adapt agricultural practices, the sector remains vulnerable, endangering both economic stability and human nutrition.